\documentclass{article}

\usepackage{arxiv}

\usepackage[utf8]{inputenc} % allow utf-8 input
\usepackage[T1]{fontenc}    % use 8-bit T1 fonts
\usepackage{hyperref}       % hyperlinks
\usepackage{url}            % simple URL typesetting
\usepackage{booktabs}       % professional-quality tables
\usepackage{amsfonts}       % blackboard math symbols
\usepackage{nicefrac}       % compact symbols for 1/2, etc.
\usepackage{microtype}      % microtypography
\usepackage{amsmath}
\usepackage{lipsum}
\usepackage{subcaption}
\usepackage{graphicx}
\usepackage{multicol,multienum, multirow}
\usepackage{threeparttable}

\title{Deep Networks with Fast Retraining}

\author{
  Wandong Zhang\thanks{The first two authors contributed equally}\\
  Electrical and Computer Engineering\\
  University of Windsor\\
  \texttt{zhang1lq@uwindsor.ca} \\
   \And
   Yimin Yang\footnotemark[1]\\
   Computer Science, Lakehead University \\
   Vector Institute for Artificial Intelligence, Toronto\\
   \texttt{yyang48@lakeheadu.ca} \\
   \And
   Jonathan Wu\thanks{Corresponding author}\\
   Electrical and Computer Engineering\\
  University of Windsor\\
  \texttt{jwu@uwindsor.ca} \\
}

\begin{document}
\maketitle

%\thanks{aaaa}

\begin{abstract}
Recent work~\cite{yang2019recomputation} has utilized Moore-Penrose (MP) inverse in deep convolutional neural network (DCNN) learning, which achieves better generalization performance over the DCNN with a stochastic gradient descent (SGD) pipeline. However, Yang's work has not gained much popularity in practice due to its high sensitivity of hyper-parameters and stringent demands of computational resources. To enhance its applicability, this paper proposes a novel MP inverse-based fast retraining strategy. In each training epoch, a random learning strategy that controls the number of convolutional layers trained in the backward pass is first utilized. Then, an MP inverse-based batch-by-batch learning strategy, which enables the network to be implemented without access to industrial-scale computational resources, is developed to refine the dense layer parameters. Experimental results empirically demonstrate that fast retraining is a unified strategy that can be used for all DCNNs. Compared to other learning strategies, the proposed learning pipeline has robustness against the hyper-parameters, and the requirement of computational resources is significantly reduced.
\end{abstract}

% keywords can be removed

\section{Introduction}
The depth of a DCNN plays a vital role in discovering intricate structures both in theory~\cite{haastad1987computational, haastad1991power, lecun2015deep, szegedy2015going} and real practice~\cite{sun2014deep, guo2014deep, sindhwani2015structured}. The original DCNN LeNet5 contains 4 convolutional layers; however, the development of computer hardware and the improvement of network topology have enabled the DCNNs to go deeper and deeper. Recent deep models, including ResNet and DenseNet, have already surpassed the 100-layer barrier at 152 and 264 layers, respectively. Although the deeper networks have strong power in discovering intricate structures in image tasks, these approaches have a key characteristic in common: All of the up-to-date DCNNs adopt SGD~\cite{bengio2007greedy, bottou2010large} and its variants\cite{kingma2014adam, duchi2011adaptive} as a foundation to train the networks. Such a training strategy makes the network suffer from being trapped in a local minimum and requiring a large amount of training time. Thus, a novel learning pipeline should be found in order to boost the generalization performance and speed up the learning.

Recent years the MP inverse technique has been utilized to train a DCNN to achieve a better generalization performance~\cite{yang2019recomputation}. Essentially, the dense layers of a DCNN (with linear activation function) can be reduced back to a linear system, whereas the approximate optimal parameters are the least squares (LS) solutions when the minimum error has been achieved. It is well established that the MP inverse, among other techniques, is the most widely known generalization of the inverse matrix to find the unique solution of an LS problem. More importantly, the work in~\cite{schmidt1992feedforward} has already proved that the output layer vector can be called the fishier vector if the weights are solved by the standard MP inverse. Following this, an increased focus has been placed on hierarchical networks with MP inverse~\cite{yang2019features, zhang2020width}. Compared to the parameters calculated with SGD, the unique solution obtained by MP inverse corresponds to the maximum likelihood estimation. To the best of our knowledge, the study in~\cite{yang2019recomputation} is the state-of-the-art work that utilizes the MP inverse in DCNN training. In each training epoch, the DCNN is first trained with the SGD optimizer; then, the parameters in dense layers are refined through the MP inverse-based approach.

Despite its advantages, the use of the training procedure for DCNN provided in~\cite{yang2019recomputation} is not as widespread as it could be. The crucial reasons for this may be as follows. On the one hand, the retraining process adds to the computational workload of each epoch. In particular, for a large dataset such as ImageNet~\cite{deng2009imagenet}, the researchers can only refine the parameters of dense layer with CPU instead of GPU acceleration because the calculation of MP inverse can occupy huge computational resources. As the training cost in the CPU environment increases dramatically, the DCNN trained with MP inverse~\cite{yang2019recomputation} is uneconomical when handling large-scale samples without access to industrial-scale computational resources.

On the other hand, before retraining the dense layer weights, the process in the study~\cite{yang2019recomputation} still requires the use of the SGD technique to optimize the parameters of all layers. According to some successful techniques~\cite{huang2016deep, gastaldi2017shake}, we hypothesize in this paper that it is not necessary for the work~\cite{yang2019recomputation} to involve the entire convolutional layer in the training process at each epoch because the refinement of parameters in dense layers can provide more clues. The existing methods~\cite{huang2016deep, gastaldi2017shake} achieved accelerating training by adjusting a set of hyperparameters on the basic SGD training pipeline. However, they do have drawbacks, such as slight degradation in performance and a lack of robustness against various environmental conditions, resulting in an unstable testing accuracy.

In this paper, we focus on providing a unified training pipeline of DCNN with better generalization performance but without incurring much additional training burden. We achieve this goal by training the DCNN with several general epochs. In each general epoch, two simple but straightforward steps that can be implemented in the pure GPU environment are employed: The first one is SGD with random learning, and the second is an MP inverse-based batch-by-batch strategy. For the first step, a freeze learning algorithm that reduces the workload and speeds up the DCNN algorithm is provided. We shorten the network by \emph{randomly} activating a $r_a$ portion of convolutional layers in each general epoch. The rate $r_a$ is preset to a value and progressively decreased. In the first several general epochs, $r_a$ is set to 1 to start the network, and all of the parameters are updated. Then, it is gradually decreased and finally comes to 0, which means that all of the convolutional layers are “frozen” without updating. Hence, the only parameter that users need to adjust is the activation rate $r_a$. As for the second one, a batch-by-batch MP inverse retraining strategy that can be processed by GPU is proposed. Instead of training all of the loaded data at once, the data is processed sequentially. By doing so, the data volume of each batch is dramatically reduced and will not consume many computational resources. Thus, the proposed training pipeline can be implemented with GPU acceleration.

In extensive experiments, several state-of-the-art deep learning architectures for pattern recognition, such as AlexNet~\cite{krizhevsky2012imagenet}, VGG-16~\cite{simonyan2014very}, Inception-v3~\cite{szegedy2016rethinking}, ResNet~\cite{he2016deep} and DenseNet~\cite{huang2017densely}, are utilized to verify the effectiveness of this method. We show across 8 datasets, including 2 large datasets (ImageNet and Place365), that the proposed method almost always improves the generalization performance without increasing the training burden. For instance, on the Food251 and Place365 datasets, fast retraining with ResNet provides 24.8\% and 21.0\% greater speeds, respectively, than the methods in~\cite{yang2019recomputation}.

\section{Related Works}
\label{sec:headings}
The training procedures of DCNN have been widely studied. However, most of these prior studies focus on improving one part of performance, either the training efficiency~\cite{hinton2012improving, huang2016deep, brock2017freezeout} or the generalization performance~\cite{yang2019recomputation, huang2017snapshot}. Few of them address both concerns.

Many successful learning schedules, such as Dropout~\cite{hinton2012improving}, Stochastic Depth~\cite{huang2016deep}, and FreezeOut~\cite{brock2017freezeout}, have already achieved a computational speedup by excluding some convolutional layers from backward pass, as the early layers of a DCNN only detect simple edge details while taking up most of the time budget. Stochastic Depth~\cite{huang2016deep} reduces the training time by removing a set of convolutional layers for each mini-batch, while FreezeOut~\cite{brock2017freezeout} reduces computational costs with cosine annealing~\cite{loshchilov2016sgdr} by freezing convolutional layers.

In study~\cite{yang2019recomputation}, a standard SGD with an MP inverse pipeline that can boost the testing performance of a DCNN was provided. It is motivated by the fact that the performance boost through network topology optimization is almost approaching \emph{its limitation}, as shown by the minimal improvement in the results of the ILSVRC competition in recent years~\cite{he2016deep, huang2017densely}. In other words, the network depth has increased dramatically recently, while the testing performance has only had a limited improvement. After each SGD training, the authors adopt the MP inverse to pull back the residual error $\textbf{e}^n$ from the output layer to each fully-connected (FC) layer in order to update the parameters. Thus, the approximate optimal parameters of FC layers can be generated. Formally, if the DCNN contains the ReLU layer and DropOut operation, the updated weight $\eta$ could be obtained via the KKT theorem~\cite{Kuhn1951Nonlinear} and the optimization solution~\cite{schmidt1992feedforward}. The parameters of the last ($n$-th) FC layer are updated through~\cite{yang2019recomputation}:
\begin{equation}
%\scriptsize
\begin{split}
&\hat{\textbf{a}}^n = \textbf{a}^n+\mu \cdot \eta^n = \textbf{a}^n+\mu \cdot \left(\left((\textbf{H}^n)^T(\textbf{H}^n) + \frac{I}{C}\right)^{-1}(\textbf{H}^n)^T\cdot \textbf{e}^n\right),
\label{1}
\end{split}
\end{equation}
where $\left((\textbf{H}^n)^T(\textbf{H}^n) + I/C\right)^{-1}(\textbf{H}^n)^T$ is the MP inverse of $\textbf{H}^n$, $\mu\in (0,1]$ is the retraining rate, $C$ is the regularization term, $\textbf{a}^n$ is the parameters of \emph{n}th FC layer, $\hat{\textbf{a}}^n$ is the updated weights. $\textbf{H}^n$ is the input feature of \emph{n}th FC layer, and $\textbf{e}^n$ is the output layer residual error.

The earlier $i$-th FC layer can be updated by:
\begin{equation}
%\scriptsize
\begin{split}
&\textbf{e}^{i} = {\rm max}\left(0, (\textbf{e}^{i+1})^T\left((\textbf{a}^{i+1})^T(\textbf{a}^{i+1})+\frac{I}{C}\right)^{-1}(\textbf{a}^{i+1})^T\right),\\
&\hat{\textbf{a}}^{i} = \mathcal F\left(\textbf{a}^{i}+\mu \cdot \left(\left((\textbf{H}^{i})^T\textbf{H}^{i} + \frac{I}{C}\right)^{-1}(\textbf{H}^{i})^T\cdot \textbf{e}^{i}\right)\right),
\label{2}
\end{split}
\end{equation}
%\vspace{-0.7cm}
where $\left((\textbf{a}^{i+1})^T(\textbf{a}^{i+1})+I/C\right)^{-1} (\textbf{a}^{i+1})^T$ and $\left((\textbf{H}^{i})^T\textbf{H}^{i} + I/C\right)^{-1}(\textbf{H}^{i})^T$ are the MP inverse of $\textbf{a}^{i+1}$ and $\textbf{H}^{i}$, respectively, $\mathcal F$ is the dropout operation, and ${\rm max}$ is the ReLU operation. After each SGD training epoch, the weights of each FC layer ${\textbf{a}}^n$ are updated.

While the method provides a strong recognition accuracy within image classification datasets, it can only be employed in a CPU environment instead of with high-speed GPU acceleration, as the parameters calculated via Eq.~\ref{1} and Eq.~\ref{2} occupy a large amount of computational resources. This work is motivated by~\cite{huang2016deep, brock2017freezeout, yang2019recomputation}, aiming at crafting a fast retraining scheme that can lead to improvements in both training speed and testing performance with all of the existing DCNN models.

\section{DCNN with Fast Retraining Strategy}

Fast retraining tunes the DCNN parameters with general epochs to achieve better generalization performance and boost the training efficiency. Each general epoch contains two steps: Step 1, the convolutional layer random learning with SGD, and Step 2, the dense layer retraining with an MP inverses-based batch-by-batch strategy.

\subsection{Step 1 - Convolutional Layer Random Learning with Stochastic Gradient Descend}

\begin{figure}[t]
\begin{subfigure}{0.49\textwidth}
   \includegraphics[trim={0.7cm, 0.7cm, 0.7cm, 0.65cm}, clip, width=\linewidth]{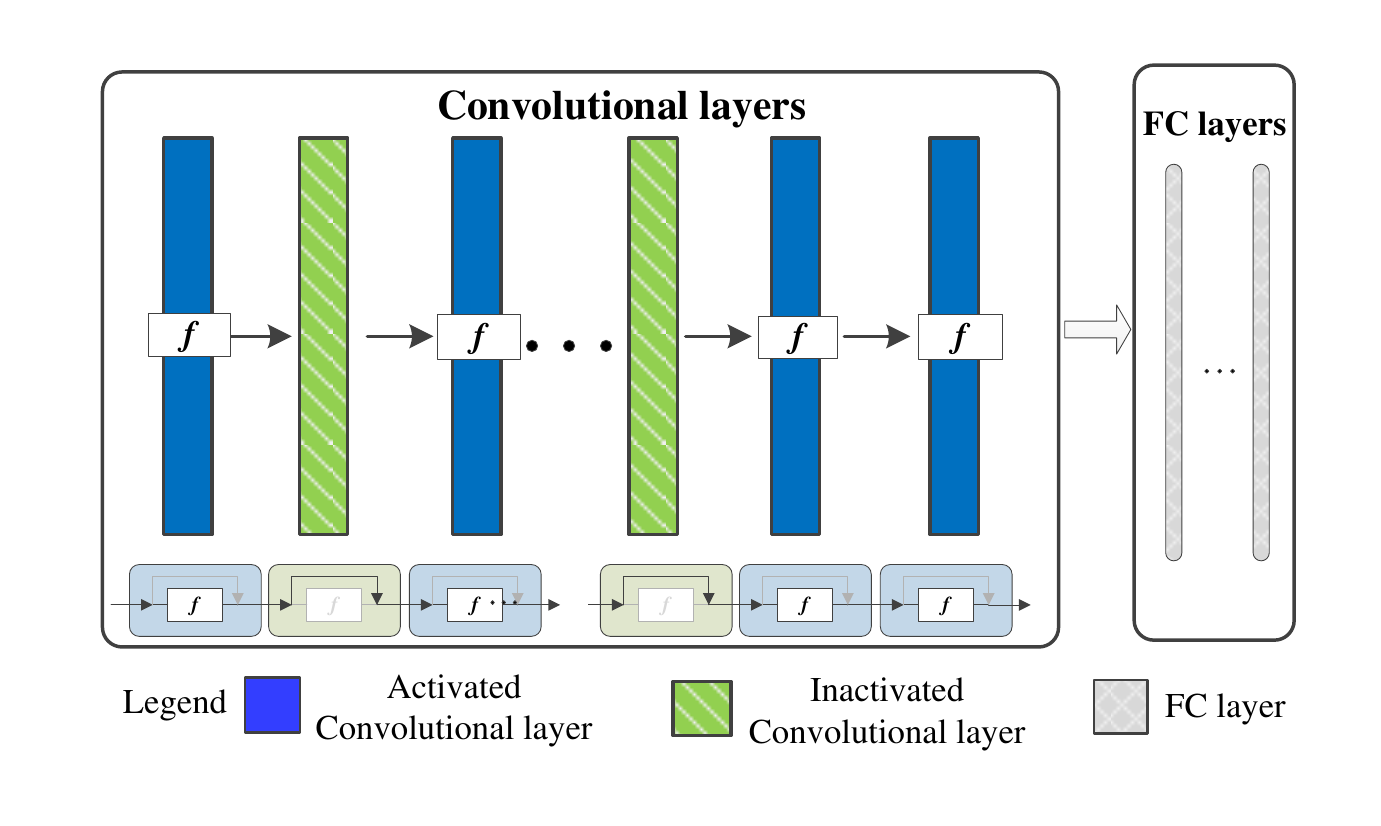}
   \caption{Step 1 - Random learning with SGD. In each epoch, users randomly activate $L_a$ number of convolutional layers, while excluding the rest $L_i$ number of convolutional layers from backward pass. $L_a$ and $L_i$ are determined by a predefined hyperparameter $r_a$.}
\label{F0002}
\end{subfigure}
\hfill
\begin{subfigure}{0.49\textwidth}
   \includegraphics[trim={0.1cm, 1.0cm, 0.15cm, 0.6cm}, clip, width=\linewidth]{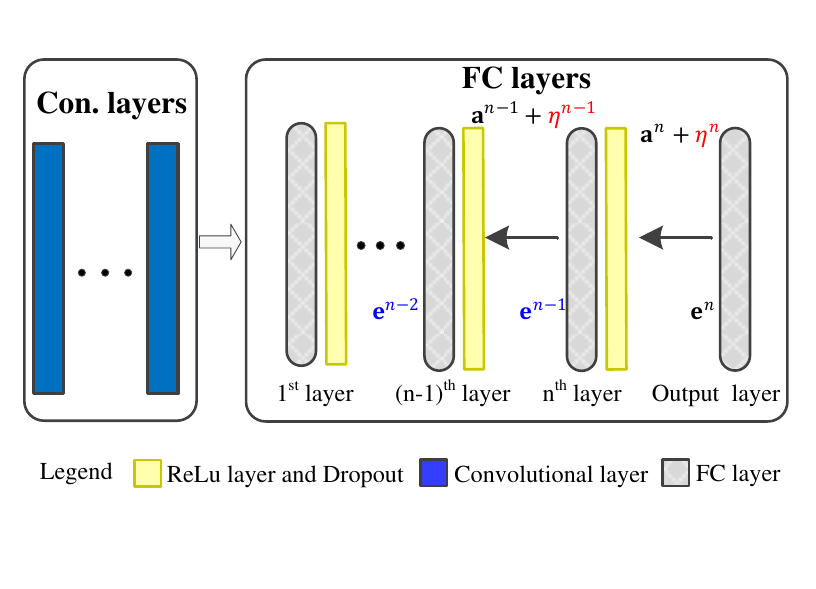}
   \caption{Step 2 - Retraining with MP inverse-based batch-by-batch strategy. $\eta^{n}$ and $\eta^{n-1}$ are obtained by Procedure I, while $\textbf e^{n-1}$ and $\textbf e^{n-2}$ are received via Procedure II. The details for Procedure I and II can be found from Algorithm 1. }
\label{F0003}
\end{subfigure}
\caption{The proposed training procedure. The DCNN is trained with general epochs containing two successive steps: Step 1 for random convolutional layer learning. Step 2 for dense layer retraining.}
\end{figure}

In this paper, we provide a simple accelerated training algorithm as depicted in Fig.~\ref{F0002} to speed up the training of a DCNN by randomly dropping the hidden layers in each epoch with a preset activation rate $r_a$. As this method only contains one hyperparameter, $r_a$, it is relatively easy for users to tune for practical employment. Note that $r_a$ keeps updating as the training epoch changes. Initially, the activation rate $r_a$ is set to 1 in the first several training epochs in order to “warm up” the DCNN network. All of the parameters in the network are tuned and updated in backward pass. After the “warm up” stage, the earlier layers are able to extract low-level features that can be used by later layers to build high-level features, and they are reliable enough to represent the raw images. Therefore, $r_a$ is increased to both accelerate network training and avoid over-fitting. In this sense, the inactivated layers are excluded from backward pass. Suppose that a designed DCNN contains $L_c$ convolutional layers; in a certain training epoch, the total number of activated ($L_a$) and inactivated ($L_i$) layers are:
%\vspace{-0.1cm}
\begin{equation}
\begin{split}
&L_a=r_a\times L_c,\,\,\text{and}\,L_i=(1-r_a)\times L_c.
\end{split}
\label{000}
\end{equation}

\subsection{Step 2 - Dense Layer Retraining with MP inverse-based Batch-by-batch Strategy}

In order to implement the retraining schedule with efficient GPUs, the feature $\textbf{H}$ and error $\textbf{e}$ in Eq.~\ref{1} is processed chunk-by-chunk with $M$ pieces, i.e., $\textbf H=\{\textbf H( \boldsymbol x_1), \textbf H( \boldsymbol x_2),\cdots, \textbf H( \boldsymbol x_M)\}$, and $\textbf e=\{\textbf e( \boldsymbol x_1), \textbf e( \boldsymbol x_2),\cdots, \textbf e( \boldsymbol x_M)\}$. First, the initial data $\textbf H( \boldsymbol x_1)$ and $\textbf e( \boldsymbol x_1)$ is given, and the weights $\eta_1$ is calculated via one-batch learning strategy~(\ref{1}). Then, the weights $\eta_i$ is updated via $\textbf H( \boldsymbol x_i)$, $\textbf e( \boldsymbol x_i)$, and $\eta_{i-1}$ in an iterative way.

Suppose we have $\textbf{H}_{p-1}, \textbf{H}_{p}, \textbf{e}_{p-1}, \textbf{e}_{p}$, which are defined as equation~(\ref{0002}).
\begin{equation}
\begin{split}
 &\textbf{e}_{p} = \begin{bmatrix}\textbf e(\boldsymbol x_1)\,\,\textbf e(\boldsymbol x_2)\,\,\cdots\,\,\textbf e(\boldsymbol x_{p})\end{bmatrix}^T = \begin{bmatrix}\textbf{e}_{p-1}\,\,\textbf e(\boldsymbol x_{p})\end{bmatrix}^T, \text{and}\\
 &\textbf{H}_{p} = \begin{bmatrix}\textbf{H}(\boldsymbol x_1)\,\,\textbf{H}(\boldsymbol x_2)\,\,\cdots\,\,\textbf{H}(\boldsymbol x_{p})\end{bmatrix}^T = \begin{bmatrix} \textbf{H}_{p-1}\,\,\textbf{H}(\boldsymbol x_{p})\end{bmatrix}^T.
 \end{split}
\label{0002}
\end{equation}

From~(\ref{1}), with $p$ batches of feature, the updated weights of one dense layer are considered as:
\begin{equation}
\eta_p=\left(\frac{I}{C}+\textbf{H}_p^T\textbf{H}_p\right)^{-1}\textbf{H}_p^T\cdot \textbf{e}_p=R_{p}^{-1}\textbf{H}_p^T\cdot \textbf{e}_p,
\label{3}
\end{equation}
where $\left(I/C+\textbf H_{p}^T \textbf H_{p}\right) \rightarrow R_{p}$. According to~(\ref{0002}), the following equation can be drawn:
\begin{equation}
\begin{split}
R_{p} = \frac{I}{C} + \begin{bmatrix}\textbf H_{p-1} \\ \textbf H(\boldsymbol x_{p})\\\end{bmatrix}^T \begin{bmatrix}\textbf H_{p-1} \\ \textbf H(\boldsymbol x_{p})\\\end{bmatrix} = R_{p-1} + \textbf H(\boldsymbol x_{p})^T \textbf H(\boldsymbol x_{p}), \\
\end{split}
\end{equation}
where $\left(I/C+\textbf H_{p-1}^T \textbf H_{p-1}\right) \rightarrow R_{p-1}$. Based on the Sherman-Morrison-Woodbury (SMW) formula~\cite{golub2012matrix}, the inverse of $R_{p}$ can be attained:
\begin{equation}
\begin{split}
R_{p}^{-1} &= \left(R_{p-1} + \textbf H(\boldsymbol x_{p})^T \textbf H(\boldsymbol x_{p})\right)^{-1}\\
&= \left(I - R_{p-1}^{-1}\textbf H(\boldsymbol x_{p})^T \left(I + \textbf H(\boldsymbol x_{p})R_{p-1}^{-1}\cdot \textbf H(\boldsymbol x_{p})^T\right)^{-1}\textbf H(\boldsymbol x_{p})\right)\cdot R_{p-1}^{-1}.
\label{003}
\end{split}
\end{equation}

The equation~(\ref{3}) can be rewritten as
\begin{equation}
\begin{split}
\eta_p=\left(I - R_{p-1}^{-1}\textbf H(\boldsymbol x_{p})^T \left(I + \textbf H(\boldsymbol x_{p})R_{p-1}^{-1}\cdot \textbf H(\boldsymbol x_{p})^T\right)^{-1}\textbf H(\boldsymbol x_{p})\right)\cdot R_{p-1}^{-1} \begin{bmatrix}\textbf H_{p-1}\\\textbf H(\boldsymbol x_{p})\\\end{bmatrix}^T \begin{bmatrix}\textbf{e}_{p-1}  \\\textbf e(\boldsymbol x_{p})\\\end{bmatrix}.
\label{004}
\end{split}
\end{equation}

Furthermore, for simplicity, we denote $K_{p}$ as:
\begin{equation}
K_{p} =  I - R_{p-1}^{-1}\textbf H(\boldsymbol x_{p})^T(\textbf H(\boldsymbol x_{p})R_{p-1}^{-1}\textbf H(\boldsymbol x_{p})^T + I)^{-1}\textbf H(\boldsymbol x_{p}).
\end{equation}

Substitute $K_{p}$ into equation~(\ref{004}), the weight $\eta_{p}$ can be simplified to the following equation:
\begin{equation}
\begin{split}
\eta_{p} = K_{p}R_{p-1}^{-1}\textbf H_{p-1}^T\textbf{e}_{p-1} + K_{p}R_{p-1}^{-1}\textbf H(\boldsymbol x_{p})^T\textbf e(\boldsymbol x_{p}).
\end{split}
\end{equation}

In the case of new training data being available, the updated weight $\eta_{p}$ can be written as:
\begin{equation}
\eta_{p} = K_{p}\eta_{p-1} + R_p^{-1}\textbf H(\boldsymbol x_{p})^T\textbf e(\boldsymbol x_{p}).
\end{equation}

\begin{figure}[t]
\vspace{-0.3 cm}
  \includegraphics[trim={0.5cm, 0.5cm, 0.8cm, 0.8cm}, clip, width=\textwidth]{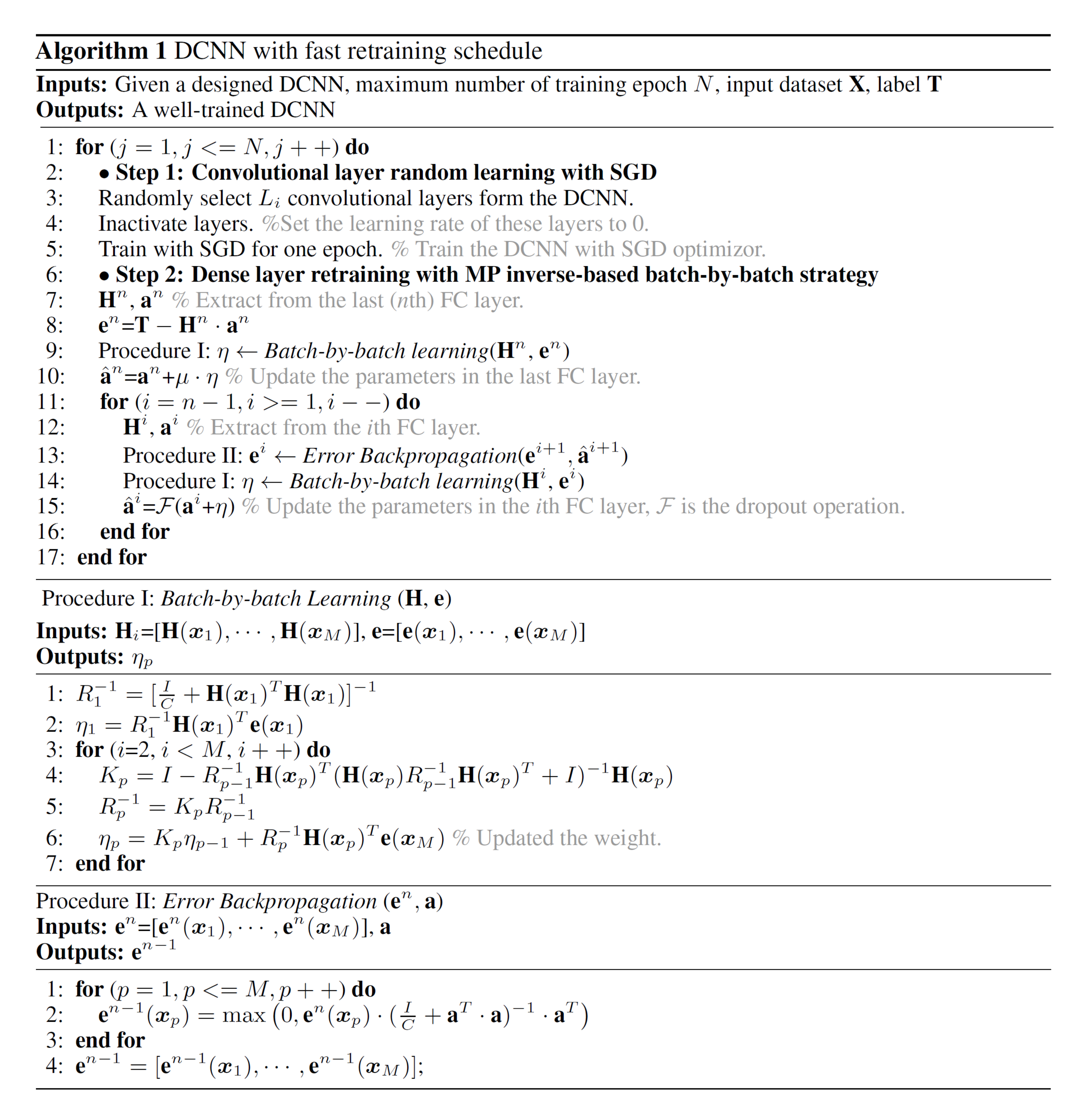}
  \label{new1}
\end{figure}

%\vspace{-0.4cm}
Above all, the parameters $\textbf{a}^n$ in the last FC layer can be updated with batch-by-batch strategy as~(\ref{et1}):
\vspace{-0.3 cm}
\begin{equation}
\begin{split}
\eta^n_p&= \begin{cases}\left((\textbf{H}^n(\boldsymbol x_1))^T\textbf{H}^n(\boldsymbol x_1) + \frac{I}{C}\right)^{-1}(\textbf{H}^n(\boldsymbol x_1))^T\cdot \textbf{e}^n, p=1\\K_{p}\eta_{p-1} + R_p^{-1}\left(\textbf H^n(\boldsymbol x_{p})\right)^T\textbf e^n(\boldsymbol x_{p}),2\leq p\leq M\\\end{cases}\\
\hat{\textbf{a}}^n&=\textbf{a}^n+\mu \cdot \eta_M^n
\label{et1}
\end{split}
\end{equation}

%\vspace{-0.6cm}
The parameters $\textbf{a}^i$ in the earlier $i$-th FC layer with $p$ batches data can be updated via~(\ref{et2}):
\begin{equation}
%\footnotesize
\begin{split}
\eta^{i}_p&= \begin{cases}\left((\textbf{H}^{i}(\boldsymbol x_1))^T\textbf{H}^{i}(\boldsymbol x_1) + \frac{I}{C}\right)^{-1}(\textbf{H}^{i}(\boldsymbol x_1))^T\cdot \textbf{e}^{i}, p=1\\K_{p}\eta^{i}_{p-1} + R_p^{-1}(\textbf H^{i}(\boldsymbol x_{p}))^T\textbf e^{i}(\boldsymbol x_{p}),2\leq p\leq M\\\end{cases}\\
\hat{\textbf{a}}^i&= \mathcal F(\textbf{a}^i+\mu \cdot\eta_M^i)
\label{et2}
\end{split}
\end{equation}
where $\mathcal F$ is the dropout operation.

%\vspace{-0.8cm}
%\subsection{Algorithmic Summary}
%\vspace{-0.2cm}

The proposed training procedure for DCNN is presented as Algorithm 1. In each general epoch, the training process can be divided into two continuous steps: Step 1, convolutional layer random learning with SGD (Line 2-5), and Step 2, dense layer retraining with an MP inverse-based batch-by-batch strategy (Line 6-16).

%-------------------------------------------------------------------------
\vspace{-0.2 cm}
\section{Experiments}
\vspace{-0.2 cm}
In this paper, we apply 8 datasets and 4 state-of-the-art DCNNs to demonstrate the efficiency of the fast retraining strategy. The experiments performed in this section were conducted on a workstation with a 256GB memory and an E5-2650 processor, and all of the DCNNs were trained on NVIDIA 1080Ti GPU.

\subsection{Dataset and Experimental Settings}

\textbf{I. Datasets.} The details of the datasets are shown as Table~\ref{T0002}. For Caltech101~\cite{fei2006one} and Caltech256~\cite{griffin2007caltech}, we randomly selected 30 images per category to get the training set, using the rest for testing. As for CIFAR100~\cite{krizhevsky2009learning}, 50,000 images were used for training and 10,000 for testing. The Food251 dataset~\cite{kaur2019foodx}, which was a food classification dataset, created in 2019, the training set (118,475 images) and validation set (11,994 images) were used for training and testing. Besides these datasets, the commonly used large-scale datasets Place365~\cite{zhou2017places} and ImageNet~\cite{deng2009imagenet} were used to evaluate the proposed work. For a comprehensive comparison, we randomly selected 200 and 500 images per category to create ImageNet-1 and Place365-1, respectively, while the validation set was used for testing.

\textbf{II. Architectures.} We evaluated fast retraining with several state-of-the-art DCNNs, such as AlexNet~\cite{krizhevsky2012imagenet}, VGG~\cite{simonyan2014very}, Inception-V3~\cite{szegedy2016rethinking}, ResNet~\cite{he2016deep}, and DenseNet~\cite{huang2017densely}. For the first three frameworks, we utilized the 16-layer VGG, the 48-layer Inception-V3, and the 50-layer ResNet, respectively. For the DenseNet, we evaluated the fast retraining scheme on two structures: the 121-layer DenseNet and the 201-layer DenseNet.

\textbf{III. Setting.} In this paper, we tested the proposed method with an original strategy under two different conditions, i.e., transfer learning and training from scratch. The experimental settings were as follows. For transfer learning, the initial learning rate was set to $1.0^{-3}$, and was divided by 10 in every 3 training epochs. The initial activation rate $r_a$ was 1, and it was set to 0.8, 0.6, and 0.4 at 25\%, 50\%, and 75\% of the total number of training epochs. Other settings include the total number of training epochs, the regularization term $C$ in retraining, and the mini batch size; these are described in Table~\ref{T0002}. As for DCNN training from scratch, we trained the model for 90 epochs. The learning rate was set to 0.1 and was lowered by 10 times at epochs 30 and 60. The $r_a$ was first set to 1 and was decreased to 0.9 and 0.6 at 50\% and 75\% of the total number of training epochs.

\begin{table}[!t]
\begin{center}
\begin{threeparttable}
\small
%\begin{tabular}{|c|r|r|r|c|c|c|c|c|c|c|c|}
\begin{tabular}{crrrcccccccc}
\toprule
\multirow{2}*{Dataset}     & \multirow{2}*{\# classes}     & \# training   & \# testing & batch size & batch size & \# max &\multirow{2}*{$C$}\\
                          &                               & samples       & samples     & (SGD)      &(MP inverse)  & epoch     &\\
\midrule
Caltech101        & 102     &3,060          &6,084      &32     &10$K$     &8/90\tnote{a}        &6/4/4/2\tnote{b}\\
Caltech256        & 257     &7,710          &22,898     &32     &10$K$     &8/90\tnote{a}        &4/4/4/2\tnote{b}\\
CIFAR100          & 100     &50,000         &10,000     &32     &10$K$     &8/90\tnote{a}        &4/2/2/2\tnote{b}\\
Food251           & 251     &118,475        &11,994     &32     &10$K$     &8/90\tnote{a}        &4/2/2/2\tnote{b}\\
Place365-1		  & 365     &182,500        &36,500     &32     &10$K$     &12/90\tnote{a}       &2/2/2/1\tnote{b}\\
Place365		  & 365     &1,803,460      &36,500     &32     &10$K$     &12/90\tnote{a}       &2/2/2/1\tnote{b}\\
ImageNet-1        & 1,000   &200,000        &50,000     &32     &10$K$     &NA/90\tnote{a}       &2/2/2/1\tnote{b}\\
ImageNet          & 1,000   &14,197,122     &50,000     &32     &10$K$     &NA/90\tnote{a}       &2/2/2/1\tnote{b}\\
\bottomrule
\end{tabular}
\begin{tablenotes}
        \scriptsize
        \vspace{0.1 cm}
        \item[a]  \noindent Configurations for transfer learning / Training from scratch  $^b$    Configurations for VGG / Inception-V3 / ResNet / DenseNet
\end{tablenotes}
\vspace{0.2 cm}
\caption{Summary of the Datasets}
\vspace{-0.4 cm}
\label{T0002}
\end{threeparttable}
\end{center}
\end{table}

\subsection{Step-by-step Quantitative Analysis}

\begin{figure}[t]
\vspace{-0.3 cm}
  \includegraphics[trim={0.5cm, 1.3cm, 0.8cm, 1.8cm}, clip, width=\textwidth]{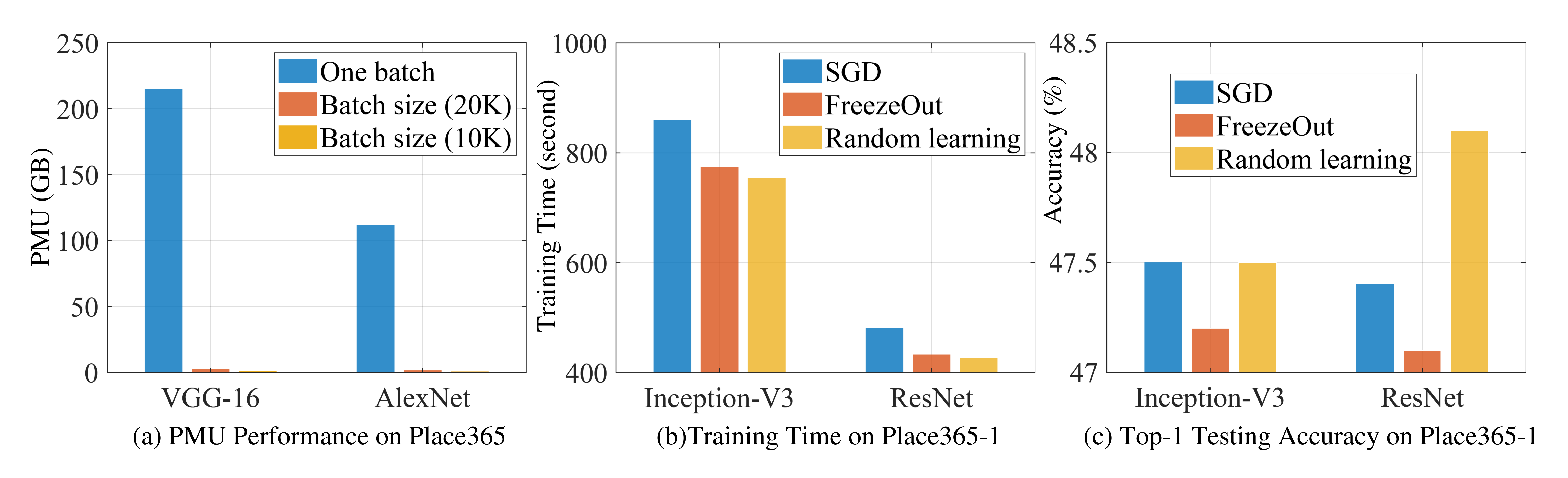}
  \vspace{-0.6 cm}
  \caption{The effective analysis results of MP inverse-based batch-by-batch strategy and convolutional layer random learning schedule. (a) The peak memory usage of one-batch and batch-by-batch on Place365 dataset. (b) Training time comparison of SGD~\cite{bottou2010large}, FreezeOut~\cite{brock2017freezeout}, and random learning on Place365-1. (c) The Rop-1 testing accuracy comparison on Place365-1 dataset.}
  \vspace{-0.4 cm}
  \label{new1}
\end{figure}

\begin{table}[!t]
\small
 \setlength\tabcolsep{5.3pt}
%\begin{tabular}{l|ccc|ccc|ccc|ccc|cccccccccccc}
\begin{tabular}{lcccccccccccccccccccccccc}
        \toprule
        \multirow{2}*{Dataset}   &\multicolumn{3}{c}{VGG-16} & \multicolumn{3}{c}{ResNet-50}  & \multicolumn{3}{c}{Inception-v3} & \multicolumn{3}{c}{DenseNet-201}      \\
        \cmidrule{2-13}
            &SGD~\cite{bottou2010large}  &R.~\cite{yang2019recomputation} &FR.   &SGD~\cite{bottou2010large}   &R.~\cite{yang2019recomputation}     &FR.    &SGD~\cite{bottou2010large}  &R.~\cite{yang2019recomputation}       &FR.    &SGD~\cite{bottou2010large}    &R.~\cite{yang2019recomputation}      &FR.   \\
         \midrule
Caltech101        &\textbf{90.4}&90.1    &90.1              &90.8    &91.3 &\textbf{91.5}     &89.7  &\textbf{90.6}   &90.5   &\textbf{92.5}  &91.7  &91.4\\
Caltech256        &69.3    &73.0    &\textbf{73.7}          &77.1    &78.2 &\textbf{78.3}     &77.9  &78.4   &\textbf{78.6}   &79.7  &81.0  &\textbf{81.2}\\
CIFAR100          &77.4    &\textbf{79.5}    &79.2          &\textbf{84.3} &83.5    &83.7     &83.8  &84.2   &\textbf{84.3}   &\textbf{86.1}  &84.4  &85.6\\
Food251           &52.7    &56.9    &\textbf{57.4}          &59.4    &61.8 &\textbf{61.8}     &59.3  &60.8   &\textbf{61.8}   &61.6  &62.4  &\textbf{62.8}\\
Place365-1        &42.9    &44.3    &\textbf{44.5}          &47.4    &48.4 &\textbf{48.6}     &47.5  &48.1   &\textbf{48.1}   &47.7  &47.7  &\textbf{47.9}\\
Place365          &50.9    &51.6    &\textbf{51.9}          &52.7    &53.3 &\textbf{53.6}     &53.8  &53.9   &\textbf{54.1 }  &54.7  &55.4  &\textbf{55.4}\\
\midrule
Average   &63.9   & 65.9&\textbf{66.2}  &68.6   &69.4  &\textbf{69.6}     &68.7   &69.4  &\textbf{69.6 }  &70.4  &70.4  &\textbf{70.7}\\
         \bottomrule
\end{tabular}
\vspace{0.1 cm}
\caption{Top-1 testing accuracy comparison (R.~\cite{yang2019recomputation} - DCNN with learning strategy in~\cite{yang2019recomputation}, FR. - DCNN with the proposed fast retraining strategy).}
\vspace{-0.4 cm}
\label{T0004}
\end{table}

In this paper, all of the experiments are evaluated with Top-1 testing accuracy, and the results recorded are the mean average of a minimum of three experiments. The best results are in boldface format.

\textbf{I. The Effectiveness Analysis of batch-by-batch strategy.} To verify the effectiveness of applying an MP inverse-based batch-by-batch scheme in dense layer retraining, we conducted a sanity check with this strategy and the one-batch schedule~\cite{yang2019recomputation}, as described in Fig.~\ref{new1}a. In particular, the peak memory usage (PMU) in training was in empirically evaluating the performance of different training modes. The investigation reveals that the batch-by-batch strategy significantly reduces the memory use of retraining DCNNs. Hence, we summarize the \textit{\textbf{first conclusion}} as such: The provided batch-by-batch method reduces the computational burden and can be accelerated in a GPU environment, which overcomes the main drawback of work~\cite{yang2019recomputation}.

\noindent \textbf{II. The Effectiveness Analysis of random learning.} To validate the effectiveness of random layer learning, experiments were conducted on the Place365-1 dataset. Note that, in this part, the retraining strategy was excluded from the training epoch (so that only random learning remains), and the DCNNs were trained on ImageNet pre-trained networks with 8 epochs. The experiments were conducted and trained under three different configurations: DCNN trained with i) the original SGD baseline, ii) the FreezeOut~\cite{brock2017freezeout} learning scheme, and iii) the proposed random learning strategy. Fig.~\ref{new1}b and \ref{new1}c compare the results. Through analysis, we reach the \textit{\textbf{second conclusion}}: The training time of the DCNN with proposed random learning is faster than that of the DCNN with traditional SGD, and it has a positive impact on generalization performance.

\noindent \textbf{III. Comparison of transfer learning.} Taking the above outcomes of sections I and II as the foundation, more experiments were carried out to compare the proposed learning procedure with the retraining algorithm~\cite{yang2019recomputation}. All of the results are tabulated in Table~\ref{T0004}. Unlike most recent works that prefer to boost the testing performance with novel network topology, in this paper, the proposed method does not contain any network modification, but it does have a slight improvement (0.1\% to 1.0\% ) on testing accuracy over the state-of-the-art MP inverse-based learning scheme~\cite{yang2019recomputation}. While the 0.1\% to 1\% Top-1 accuracy boost seems to lead to marginal improvement, it is not easy to obtain these improvements at the current stage as the DCNN optimization is almost achieving its limitation. For example, VGG-16 and ResNet are the ILSVRC winners in the year of 2014 and 2016, respectively. ResNet only provides 1.2\% boost than VGG-16 on CIFAR100 set, but 1\% lower on SUN397 set.

\begin{figure*}[t]
\begin{center}
  \includegraphics[trim={1.5cm, 1.5cm, 1.5cm, 1.5cm}, clip, width=1\linewidth]{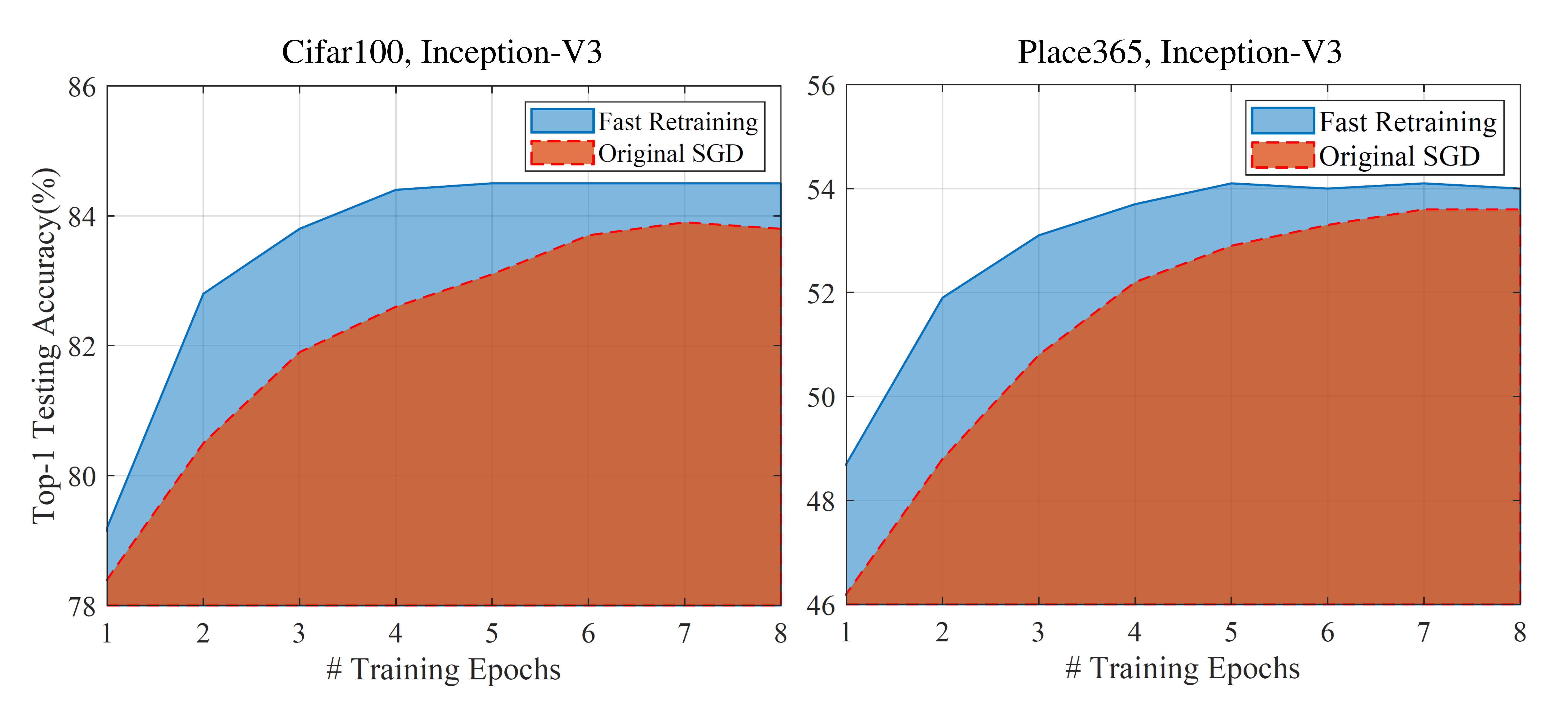}
\end{center}
\vspace{-0.3 cm}
  \caption{Top-1 testing accuracy of InceptionNet on CIFAR100 and Place365-1 datasets.}
\label{F0006}
\end{figure*}

Furthermore, the total training time of the fast retraining method, the recomputation method~\cite{yang2019recomputation}, and the original SGD method are tabulated in Table~\ref{T0005}. Note that all of the experiments were conducted with 8 training epochs. Figure~\ref{F0006} shows the generalization performance on these datasets by a plot as the number of training epochs increases. We can easily find that: The proposed strategy presents a speedup of up to 25\% compared to the existing retraining strategy~\cite{yang2019recomputation}; It only needs 3 to 4 epochs to get the optimal results, whereas the original DCNN needs at least 6 epochs. Through Table~\ref{T0004}, Table~\ref{T0005}, and Fig.~\ref{F0006}, the \emph{\textbf{last conclusion}} that can be drawn is: The fast retraining scheme improves the generalization performance of a DCNN, reducing the learning time by 15\% to 25\% over the existing MP inverse-based learning paradigm~\cite{yang2019recomputation}.

\noindent \textbf{IV. Comparison results of training from scratch.} In order to extensively test the fast retraining method, we employed another set of experiments under the condition of training from scratch. Table~\ref{T0006} shows the comparison results with InceptionNet and DenseNet on the ImageNet-1 and ImageNet datasets. Through Table~\ref{T0006}, we find that the Inception-v3 and DenseNet-121 with fast retraining could have 1.9\% and 0.9\% improvement over those with the traditional SGD scheme, and there is a 0.6\% and 0.3\% boost over the training pipeline in~\cite{yang2019recomputation}. Thus, the effectiveness of the proposed fast retraining is verified.

\begin{table}[!t]
\begin{center}
\small
%\begin{tabular}{|c|c|r|r|r|r|}
\begin{tabular}{ccccccc}
\toprule
Dataset          & DCNN           &SGD~\cite{bottou2010large}      & Retraining~\cite{yang2019recomputation}       & Fast Retraining  & Imp. - SGD (\%)     & Imp. - R. (\%)\\
\midrule
\multirow{2}*{CIFAR100}         &Inception                         &282     &308   &262   &8.7     &15.0\\
                                 &ResNet                           &161     &176   &139   &\textbf{16.3}    &21.0\\
\multirow{2}*{Place365-1}         &Inception                       &860     &968   &772   &12.1    &20.2\\
                                     &ResNet                       &481     &589   &445   &10.8    &\textbf{24.8}\\
\bottomrule
\end{tabular}
\end{center}

\caption{Comparison of total training time with SGD, retraining strategy and the proposed fast retraining on CIFAR100 and Place365-1 datasets in \emph{minute} (Imp. - SGD (\%): the improvement over SGD optimizer, Imp. - R. (\%): the improvement over retraining schedule).}
\label{T0005}
\end{table}

\begin{table}[!t]
\begin{center}
\small
\vspace{-0.1 cm}
%\begin{tabular}{|c|c|c|c|c|c|}
\begin{tabular}{cccccc}
\toprule
Method          & Dataset           &Accuracy (\%) \\
\midrule
%ResNet-50 with Orig.      &ImageNet-1                         &46.8\\
Inception-v3 with SGD~\cite{bottou2010large}.    &ImageNet-1                         &42.2\\
DenseNet-121 with SGD~\cite{bottou2010large}.       &ImageNet                           &69.1\\
\midrule
%ResNet-50 with R.       &ImageNet-1         &47.7\\
Inception-v3 with retraining scheme~\cite{yang2019recomputation}   &ImageNet-1         &43.5\\
DenseNet-121 with retraining scheme~\cite{yang2019recomputation}      &ImageNet        &69.9\\
\midrule
%ResNet-50 with FR.       &ImageNet-1         &\textbf{48.2}\\
Inception-v3 with fast retraining scheme    &ImageNet-1         &\textbf{44.1}\\
DenseNet-121 with fast retraining scheme       &ImageNet           &\textbf{70.2}\\
\bottomrule
\end{tabular}
\end{center}
\caption{Top-1 testing accuracy comparison under the condition of training from scratch.}
\vspace{-0.3 cm}
\label{T0006}
\end{table}

\vspace{-0.2 cm}
\section{Conclusion}
\vspace{-0.2 cm}
In this paper a unified fast retraining procedure for DCNN is proposed. Compared to the state-of-the-art DCNN training strategy~\cite{yang2019recomputation}, this method achieves better testing performance but without occupying much computation resources. In particular, it provides a random learning schedule to speed up the convolutional layer learning and a batch-by-batch Moore-Penrose inverse-based retraining strategy to optimize the parameters of dense layer. This scheme can be applied to all DCNNs, and the batch-by-batch solution of Moore-Penrose inverse allows the proposed training pipeline to be accelerated in a pure GPU environment. The experimental results on benchmark datasets prove the effectiveness and efficiency of the proposed fast retraining algorithm.

\bibliographystyle{plain}
\bibliography{example_paper}

\end{document}